\documentclass[conference]{IEEEtran}
\usepackage{times}
\usepackage{helvet}
\usepackage{courier}
\usepackage{graphicx}
\usepackage{caption}
\usepackage{subcaption}
\usepackage{epstopdf} 
\usepackage{booktabs}
\usepackage[noend]{algpseudocode}
\usepackage{verbatim}
\usepackage[cmex10]{amsmath}
\usepackage{url}
\usepackage{xcolor}
\usepackage[normalem]{ulem}
\useunder{\uline}{\ulined}{}
\DeclareUrlCommand{\bulurl}{}

\makeatletter
\def\BState{\State\hskip-\ALG@thistlm}
\makeatother

\begin{document}

%+++++++++++++++++++++++++++++++++++++++++++
\title{LIDE: Language Identification from Text Documents}

\author{\IEEEauthorblockN{Priyank Mathur}
\IEEEauthorblockA{
Adobe Systems\\
Stanford University\\
Stanford, California 94305\\
priyank.mathur [at] gmail.com}
\and
\IEEEauthorblockN{Arkajyoti Misra}
\IEEEauthorblockA{
Target Corporation\\
Stanford University\\
Stanford, California 94305\\
arkajyoti [at] gmail.com}
\and
\IEEEauthorblockN{Emrah Budur}
\IEEEauthorblockA{
Garanti Technology\\
Stanford University\\
Stanford, California 94305\\
emrah.budur [at] yahoo.com}}

\date{}

\maketitle

\begin{abstract}
The increase in the use of microblogging came along with the rapid growth on short linguistic data.  On the other hand deep learning is considered to be the new frontier to extract meaningful information out of large amount of raw data in an automated manner.  In this study, we engaged these two emerging fields to come up with a robust language identifier on demand, namely \textit{Language Identification Engine (LIDE)}.  As a result, we achieved 95.12\% accuracy in \textit{Discriminating between Similar Languages (DSL) Shared Task 2015} dataset, which is comparable to the maximum reported accuracy of 95.54\% achieved so far.  
\end{abstract}

\section{Introduction}
 
Automatic language detection is the first step toward achieving a variety of tasks like detecting the source language for machine translation, improving the search relevancy by personalizing the search results according to the query language  \cite{Stiller2010AmbiguityDetection,Datta}, providing uniform search box for a multilingual dictionary \cite{Nguyen2013WordCommunication}, tagging data stream from Twitter with appropriate language etc. While classifying languages belonging to disjoint groups is not hard, disambiguation of languages originating from the same source and dialects still pose a considerable challenge in the area of natural language processing. Regular classifiers based on word frequency only are inadequate in making a correct prediction for such similar languages and utilization of state of the art machine learning tools to capture the structure of the language has become necessary to boost the classifier performance. In this work we took advantage of recent advancement of deep neural network based models showing stellar performance in many natural language processing tasks to build a state of the art language classifier.

We benchmarked our solution with the industry leaders and achieved near perfect score in the DSL test dataset.
\section{Previous Work}

In the past, a variety of methods have been tried like Naive Bayes \cite{Lui2012}, SVM  \cite{Bhargava2010LanguageSVMs}, n-gram \cite{Cavnar1994}, graph-based n-gram \cite{Tromp2011}, prediction partial matching (PPM)  \cite{Bobicev2013NativePPM}, linear interpolation with post independent weight optimization and majority voting for combining multiple classifiers  \cite{Carter2012} etc. and the best accuracy achieved are still in the lower ninety percents.  

The researchers have worked on various critical tasks challenging the dimensions of the topic, including but not limited to, supporting low resource languages, i.e. Nepali, Urdu, and Icelandic \cite{Bergsma2012,Semicoast2010} 
handling user-generated unstructured short texts, i.e. microblogs \cite{Bergsma2012,Carter2012}
building a domain agnostic engine \cite{Bergsma2012,Tromp2011}.
Existing benchmarking solutions approach the LID problem in different ways where LogR \cite{Bergsma2012} adopts a discriminative approaches with regularized logistic regression, TextCat and Google CLD \cite{Sites2013} recruits N-gram-based algorithm,  langid.py \cite{Lui2012} relies on a Naive Bayes classifier with a multinomial event model.

The outstanding results, of the time, suggested by Cavnar and Trenkle became de facto standard of LID even today \cite{Cavnar1994}.  The significant ingredient of their method is shown to use a rank order statistic called "out of place" distance measure \cite{Dunning1994StatisticalLanguage}.  
\iffalse
Basically, they profiled each language and query as a list of its most frequent n-grams in descending order.  The total distance of ranks of n-grams between a particular language profile and query profile is considered to be the distance of query to that particular language.  As a result, the closest language is predicted as the language of the query.  
\fi 
The problem in their approach is that they generated n-grams out of words that requires tokenization. However, many languages including Japanese and Chinese have no word boundaries.  Considering that Japanese is the second most frequent language used in Twitter  \cite{Semicoast2010}, there is a need for better approach to scale the solution to all languages. As a solution to their problem, Dunning came up with a better approach with incorporating byte level n-grams of the whole string instead of char level n-grams of the words  \cite{Dunning1994StatisticalLanguage}.  

\iffalse
The laborious nature of the data labeling limited the size of training set and the number of language  \cite{Tromp2011,Carter2012}. However, it was shown to be near-perfectly resolvable by incorporating scalable workforce, i.e. Amazon Mechanical Turk and systematic identification, i.e. filtering out tweets belonging to followers of language-specific sources, i.e. BBC-Urdu \cite{Bergsma2012}.  
\fi
\iffalse
In addition, including additional metadata, such as 
geolocation of tweets  \cite{Bergsma2012}, 
the links referring to content elsewhere providing additional textual information \cite{Carter2012},
the language characteristic of hashtags, usernames, conversation history
was proven to be effective for increasing the accuracy of the engine.  However, incorporation of additional data limit the scalability of the solution into multiple domains which do not have those information available in their context.
\fi
After a rigorous literature survey, we found no prior study that applied deep learning on language identification of text.  On the other hand, there are a few number of studies that applied deep learning to identify the language of speech  \cite{Lopez-Moreno2014AutomaticNetworks,Montavon2009DeepIdentification,Jiang2014DeepIdentification,Gonzalez-Dominguez2014AutomaticNetworks}. We believe this study will be the first in the literature if published for LID in textual data by means of deep learning.

\section{Dataset description}

The data for this project work was obtained from "Discriminating between Similar Language (DSL) Shared Task 2015" \cite{LT4VarDialWorkshop}. A set of 20000 instances per language (18000 training (train.txt) and 2000 evaluation (test.txt)) was provided for 13 different world languages. The dataset also consisted of a subset (devel.txt) of the overall training data which we utilized for hyper-parameter tuning. The languages are grouped as shown in Table \ref{table:dataset_description_table}.  The names of the groups will be frequently referred in the subsequent sections. 
\begin{table}[h]
\centering
\resizebox{\columnwidth}{!}{%
\begin{tabular}{@{}llrr@{}}
\toprule
\toprule
Group Name & Language Name &  Language Code \\ 
\midrule
\midrule
South Eastern Slavic & Bulgarian & bg\\
&Macedonian & mk \\
\midrule
South Western Slavic & Bosnian &  bs \\
 & Croatian &  hr \\
 & Serbian  &  sr \\
\midrule
West-Slavic & Czech & cz \\
& Slovak & sk \\
\midrule
Ibero-Romance (Spanish) & Peninsular Spain & es-ES \\
& Argentinian Spanish & es-AR \\
\midrule
Ibero-Romance (Portuguese) & Brazilian Portuguese & pt-BR\\
& European Portuguese & pt-PT \\
\midrule
Astronesian & Indonesian & id \\
 & Malay & my \\
\bottomrule
\bottomrule
\end{tabular}
}
\vspace{0.2cm}
\caption{Benchmark results of available solutions}
\label{table:dataset_description_table}
\end{table}

Each entry in the dataset is a full sentence extracted from journalistic corpora and written in one of the languages and tagged with the language group and country of origin. A similar set of mixed language instance was also provided to add noise to the data. A separate gold test data was provided for the final evaluation (test-gold.txt). 

\begin{figure}
\centering
\resizebox{\columnwidth}{!}{
\begin{tabular}{cc}
\includegraphics[width=40mm]{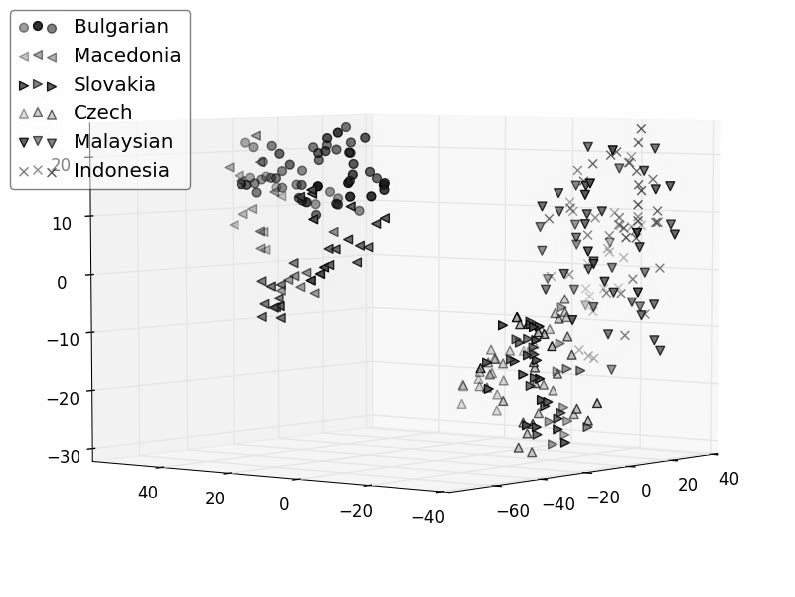}
\label{fig:less_sim}
&
\includegraphics[width=40mm]{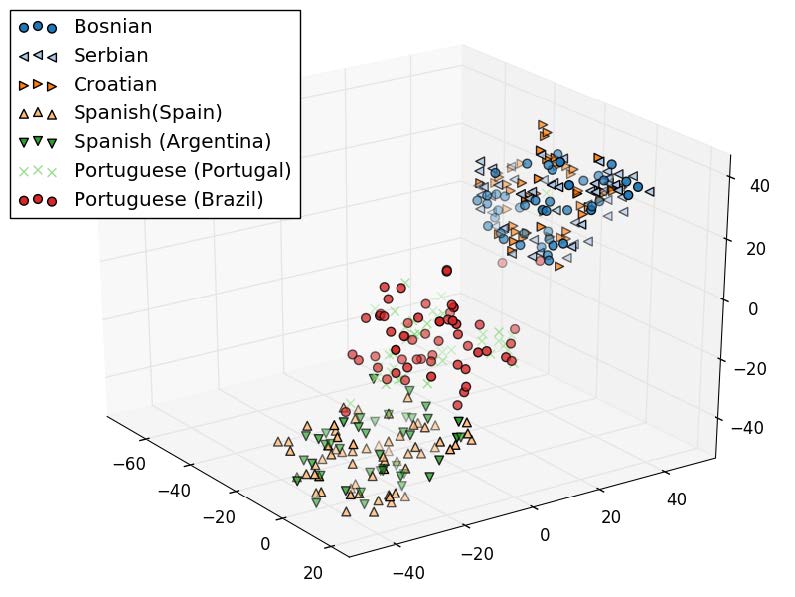}
  \label{fig:more_sim}\\
\small (A) Easily separable &  (B) Difficult to separate \\
\end{tabular}
}
\caption{t-SNE visualization of language groups. More plots including 3D animated plot are available at : \textit{http://SeeYourLanguage.info}}
\label{fig:language_groups_visualization}
\end{figure}

We applied t-SNE algorithm to visualize the instances in 3D euclidean space \cite{T-SNEMaaten,VanDerMaaten2014AcceleratingAlgorithms}.  For feature extraction, we vectorized each sentence over 1 to 5-grams of the tokens delimited by white space characters.  Fig. \ref{fig:language_groups_visualization} shows the resulting plot. As can be seen on the plot, the languages in the same group overlap a lot while the languages in different groups can be linearly separable. A 3 dimensional visualization of all the languages can be viewed at \bulurl{www.youtube.com/watch?v=mhRdfC26q78}.

\section{Methods}

\subsection{Multinomial Naive Bayes}
We created a baseline result by training a Multinomial Naive Bayes model because it is quick to prototype, runs fast and known to provide decent results in the field of text processing. We have done no pre-processing of the text commonly done in the field like stemming or stop word removal because we believe that could potentially remove important signatures of a particular language, particularly when the same language is spoken by two geographically disconnected group of people (e.g Portuguese spoken in Portugal and Brazil). We experimented with both word and character n-grams. The character n-grams turned out to be particularly useful when differentiating between two languages using mostly distinct character sequences in their alphabet. 

\begin{figure}[h]
\centering
\includegraphics[width=3.5in]{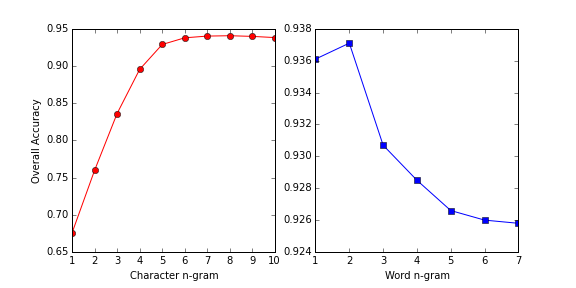}
\caption{Naive Bayes performance as a function of n for both word and character n-grams.}
\label{fig:char_vs_word}
\end{figure}

The character level n-gram behaves quite differently from that of word level n-grams as shown in Fig. \ref{fig:char_vs_word}. Single characters carry little information and therefore the performance for character n-gram improves quite sharply as the number of characters is increased before saturating at about n=8. 
We experimented with character n-grams both restricted at word boundaries and spanning across word boundaries. The latter has a marginal performance boost at the cost of longer training time and memory pressure.

The word n-gram model peaks at n=2 and drops beyond that. While higher order n-grams carry more structure of the language, they become increasingly infrequent too and therefore the models don't always get a boost from it. Both the character level and word level n-gram models show similar performance where they really excel at certain languages (Czech, Slovak) and do poorly at other (Bosnian, Croatian, Serbian). 

\subsection{Logistic Regression}

\begin{figure}
\centering
\begin{tabular}{cc}
\includegraphics[width=40mm]{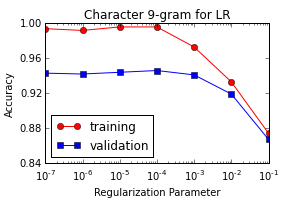}
&
\includegraphics[width=40mm]{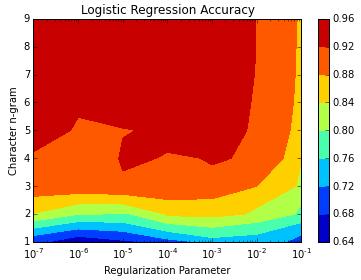}

\end{tabular}

\caption{The LR model was able to completely fit the training data but the accuracy on validation data peaked at about 94.5\% overall.}
\label{train_valid}
\end{figure}

We next tried a regularized logistic regression and here too the character level n-gram performed a little better than the word n-grams. Fig. \ref{train_valid} shows that the model was able to completely fit the training set but the performance on the validation set plateaued close to 0.945. The best performance was obtained by a character 9-gram model that includes all n-grams up to n=9. These n-grams were truncated at the word boundaries, or in other words these n-grams did not capture two or more consecutive words. Relaxing this criterion significantly increases the size of the term frequency matrix and pushes the boundary of the computer memory but it does improve the performance by a fraction of a percent.

\subsection{Recurrent Neural Network}

The MNB and LR approaches work really well in distinguishing two languages that have very little in common because the set of n-grams will have very little overlap between them. This approach does not work very well when two languages are close to each other and share a lot of words between them. Therefore, it becomes necessary to capture the structure of a languages better to distinguish between similar languages. We explored Recurrent Neural Networks (RNN) for this purpose.

RNNs are a special kind of neural networks which possess an internal state by virtue of a cycle in their hidden units. As such, RNNs are able to record temporal dependencies among the input sequence, as opposed to most other machine learning algorithms where the inputs are considered independent of each other. Hence, they are very well suited to natural language processing tasks and have been successfully used for applications like speech recognition, hand writing recognition etc.

\begin{figure}[h]
\centering
\includegraphics[scale=0.2]{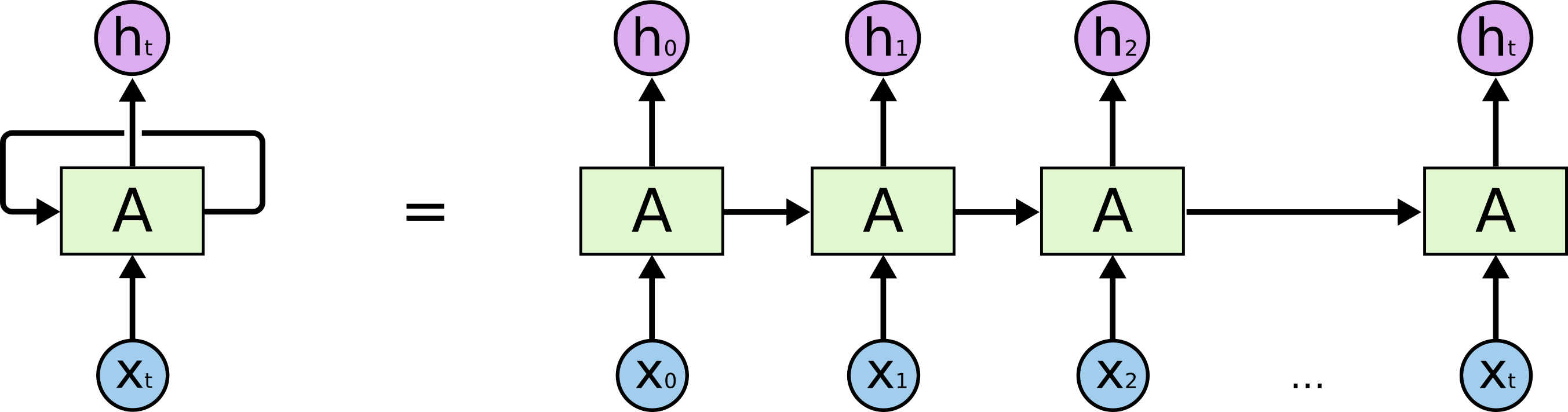}
\caption{Visualization of an un-rolled recurrent neural network \cite{UnderstandingBlog}}
\label{fig:unrolled_rnn}
\end{figure}

Until recently, RNNs were considered very difficult to train because of the problem of exploding or vanishing gradients \cite{Pascanu2013OnNetworks}  which makes it very difficult for them to learn long sequences of input. Few methods like gradient clipping have been proposed to remedy this. Recent architectures like Long Short Term Memory (LSTM)  \cite{Schmidhuber1997LongMemory} and Gated Recurrent Unit (GRU)  \cite{Cho2014LearningTranslation} were also specifically designed to get around this problem. In our experiments, we used single hidden layer recurrent neural networks that used gated recurrent units. \\

\subsubsection*{Hyper-parameter tuning}
In our single layer networks, we had three model hyper parameters to search over 
\begin{enumerate} 
 \setlength{\itemsep}{-2ex}  
 \setlength{\parskip}{0ex} 
 \setlength{\parsep}{0ex}
\item Epochs - the number of iterations over training data. We generally try to train until the network saturates.\hfil\break
\item Hidden layer size - Number of hidden units in the hidden layer.\hfil\break 
\item Dropout - Deep neural networks with large number of parameters are very powerful machines but are extremely susceptible to overfitting. Dropout provides a simple way to remedy this problem by randomly dropping hidden units as each example propagates through the network and back  \cite{Srivastava2014Dropout:Overfitting}.\hfil\break 
\end{enumerate}

We used a subset of our overall training data (devel.txt) for hyper parameter selection. This subset was further divided into 75\% training data and 25\% validation data. In the first step of the process, we varied a single parameter while keeping the other two constant. The plots below (fig. 4) show the performance of the resultant models on the validation dataset as each parameter was changed.

\begin{figure}[h]
\centering
\includegraphics[scale=0.2]{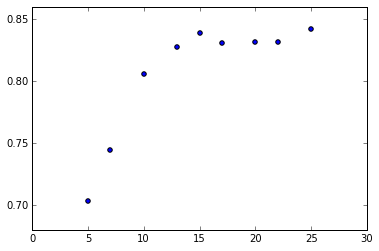}
\includegraphics[scale=0.2]{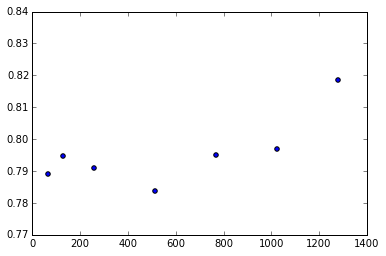}
\includegraphics[scale=0.2]{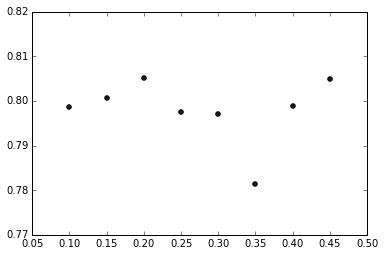}
\caption{Variation of the accuracy on validation dataset as we vary training epochs, number of hidden units and drop off }
\label{fig:acc_vs_params}
\end{figure}

\begin{figure}[h]
\centering
\includegraphics[scale=0.25]{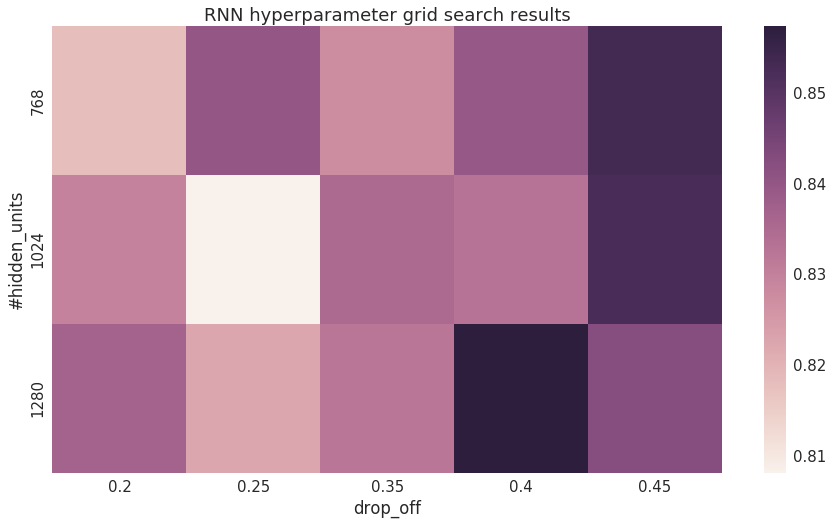}
\caption{Grid search over the best parameter values found in the previous step}
\label{fig:grid_search}
\end{figure}

As we can see in the plot above in Fig.\ref{fig:acc_vs_params}, increasing the number of training epochs improves the model performance up to a certain stage, after which it plateaus. Hence, for the next stage of tuning, we fixed the number of training epochs to 20. Using the best values for the number of hidden units and dropout found above, we performed grid search over all combinations of these parameters. The result of the grid search is visualized in Fig. \ref{fig:grid_search}. The (number of hidden units, dropout) combinations (1280, 0.4) and (768, 0.45) gave us the best performance on the validation set. The final values chosen for further experimentation were 768 hidden units and 0.45 dropout so as to avoid overfitting.\\

\subsubsection*{Training procedure}
Our final model is an ensemble of 5 RNNs, each built using a different feature set, namely, from character 2-grams to character 5-grams and word unigrams. To train our models, we divided our entire training data (train.txt) into 90\% training set and 10\% validation set. Once trained, we measured the performance of each model individually on the validation set and is reported in Table \ref{table:performance}.

\begin{figure}[h]
\centering
\includegraphics[scale=0.4]{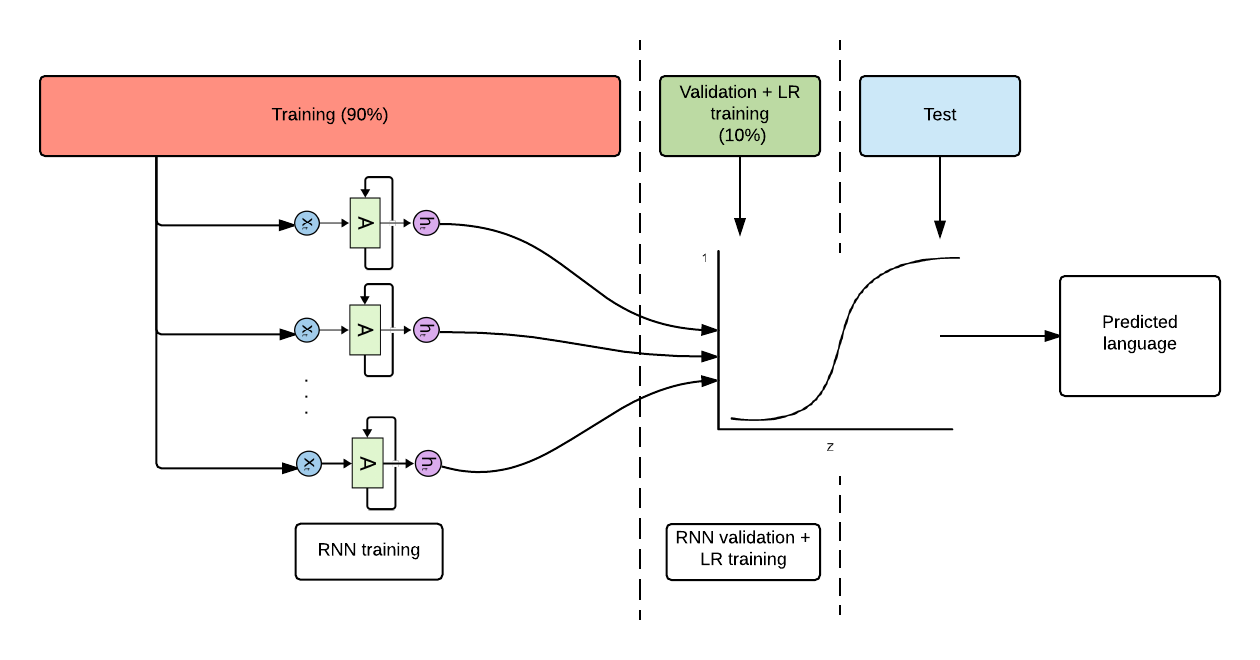}
\caption{Training procedure for the RNN ensemble}
\label{fig:rnn_training}
\end{figure}

As seen in Fig. \ref{fig:rnn_training}, to construct the ensemble, instead of manually assigning weights to each model, we constructed a Logistic Regression model to get the final output. The features for this LR model were the outputs from the 5 RNNs created earlier and it was tuned using 5 fold cross validation over the 10\% validation dataset. 

For training the RNNs, we used a Python library called \textit{passage}, which is built on top of Theano. Although the library provides several tools for text pre-processing including tokenization, it lacked the ability to generate character n-gram level features. Therefore, we had to extend the library with custom character level feature generators. In addition, training neural networks on CPUs consumes a lot of time. Hence, for our experiments, we leveraged AWS GPU (g2.2xlarge) instances that provided a 10x boost in time required to train one model.

\section{Results}

Table \ref{table:performance} shows a comparison of the models we have experimented with. One surprising feature of the result is that individual RNN models were not able to beat the performance of the MNB and LR models, even though the latter models have minimal knowledge of a language structure. However, when we created an ensemble of RNN models, it turned out to be the best model and crossed the 95\% threshold for the first time. It should be noted that for a particular $n$-gram model, MNB and LR models use all $m$-grams where $1 \leq m \leq n$. However, due to the very nature of an RNN architecture, a combination of n-grams cannot be used because that will lead to an overlapping sequence of  content to be fed to the network. Since any given n-gram captures only limited information about a language, it was natural to try an ensemble of $n$-gram RNN models with different values of $n$, so that structure of the language can be captured at multiple different levels. 

The boost in performance due to ensemble can also be attributed to model combination, which aims to achieve at least as good of a performance  as the worst model in the ensemble. This is because  individual models can make mistakes on different examples, and therefore, by using an ensemble we are able to reduce this variance. While we tried other model combination strategies like median and manual weighting, building a Logistic Regression classifier on top of RNNs really helped us find the optimal weight that should be given to each individual model.
We could not include RNN models beyond character 5-gram in the ensemble because of memory limitation and including the MNB or LR model in the ensemble did not improve the performance of the model.

\begin{table}
\centering
\resizebox{\columnwidth}{!}{%
\begin{tabular}{@{}lcc@{}}
\toprule
\toprule
 &  \multicolumn{2}{c}{Accuracy} \\
Model & Validation Set &  Test Set  \\
\toprule
\toprule
MNB (char 9-gram) & 0.9479 & 0.9452 \\
LR (char 9-gram) & 0.9486 & 0.9449 \\
RNN (char 2-gram) & 0.9200 & 0.9213 \\
RNN (char 3-gram) &	0.9328 &	0.9338 \\
RNN (char 4-gram) &	0.9377 &	0.9347 \\
RNN (char 5-gram) &	0.9347 &	0.9316 \\
RNN (word uni-gram) &	0.9351 &	0.9330 \\
Ensemble of RNN model (LIDE) &	0.9533	& 0.9512 \\
\bottomrule
\end{tabular}
}
\vspace{0.2cm}
\caption{Performance comparison of various models}
\label{table:performance}
\end{table}

\section{Discussion}
\label{sec:disc}

\begin{figure}[h]
\centering
\resizebox{\columnwidth}{!}{%
\includegraphics[scale=0.38]{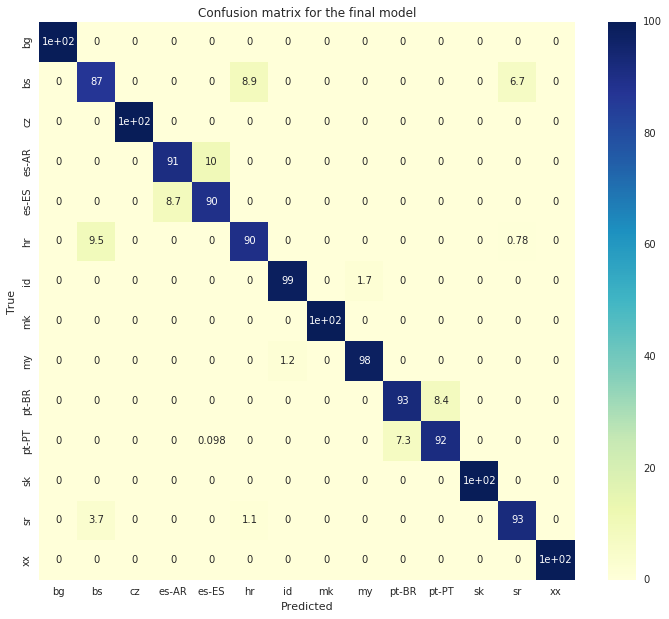}
}
\caption{Confusion matrix}
\label{fig:conf_matrx}
\end{figure}

The final classification for each language group is captured in the confusion matrix in Fig. \ref{fig:conf_matrx}. It is quite evident from our results that the biggest challenge consistently posed to our classifiers is distinguishing the languages in South Western Slavic group (\textit{bs, hr, cr}). The training set revealed that among all the words in \textit{bs}, 48\% are common to \textit{hr} and 41\% to \textit{sr}. Since Fig. \ref{train_valid} clearly showed we didn't underfit the training set, it made sense to augment the training data in these three language categories. We incorporated a significantly larger labeled data for two of these languages and also downloaded newspaper articles in \textit{bs}, but the classification accuracy in this language group did not improve. Looking closer to some of these external datasets revealed that none of the new words could be uniquely associated to any of the three languages and therefore, the additional data probably added more noise than signal.

\begin{figure}
\centering
\resizebox{\columnwidth}{!}{%
\begin{tabular}{cc}
%\hline
\includegraphics[width=40mm]{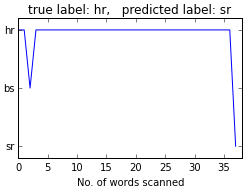}
&
\includegraphics[width=40mm]{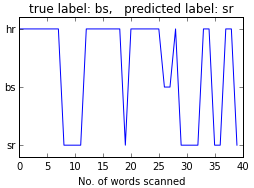}
\\
\includegraphics[width=40mm]{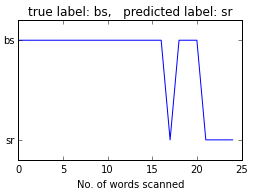}
&
\includegraphics[width=40mm]{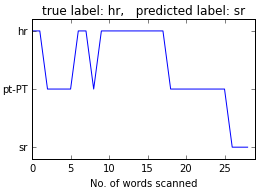}

\end{tabular}
}
\caption{Scenarios where the LR classifier incorrectly predicted the target language. (Detailed description in Section \ref{sec:disc})}
\label{failure}
\end{figure}

To understand the failure mechanism of the classifier for the South Western Slavic language group, we fed the LR classifier, which is the best of single models in validation set according to Table \ref{table:performance},  different fractions of a document it failed to classify correctly. For example, the following document is in Bosnian(\textit{bs}) but the classifier predicts its language as Serbian(\textit{sr}): 
\textit{Usto se osvrnuo na ekonomsku situaciju u kojoj je veliki broj novinara u potrazi za poslom, na mizerne plaće i guranje etičke strane profesije u zapećak}. So we fed the classifier with "\textit{Usto}" and noted the prediction, then fed it with "\textit{Usto se}" and noted the prediction, and so on until the full sentence is fed. The classifier prediction at different stages of the sentence scan is plotted in Fig.\ref{failure}. The top left panel of Fig.\ref{failure} shows that the classifier for the most part thinks the document to be actually \textit{bs}, until it saw the last word of the sentence when it switched its prediction to \textit{sr}. We think this is due to the fact that the last word associated very uniquely to \textit{sr} in the training corpus. The bottom left panel of Fig.\ref{failure} shows a similar scenario but in this case the classifier switched back and forth a couple of times. The 'confusion' of the classifier is very high in the top right panel of the figure because the particular sentence was made of words and phrases that are common to all three languages. We believe that the correct classification of such documents needs creation of extra features based on deeper understanding of this language group. Another possible scenario where any classifier can struggle is when the body of the text contains a quotation of a different language. The bottom right panel of  Fig.\ref{failure} shows a scenario where a document in Serbian had a comment in Portuguese, though that was not the cause of the eventual classification failure. Removing quotes from a document is a potential option but it can also have adverse effect if the quote is in the same language as that of the main document.

\section*{Comparison with other systems}
We assessed the performance of LIDE by comparing its result with the domain leaders in an \textit{unfair} test  described below.\\ 
We queried the \textit{test file} of dataset of DSL Shared Task 2015  and accepted the resulting predictions \textit{even if} 
\begin{itemize}
\item the dialect of the language is not distinguished in Ibero-Romance language group due to lack of support, i.e. Google always predicts \textit{Portuguese} for sentences both in Brazilian Portuguese and European Portuguese.
\item a certain language is not supported at all, i.e. Rosette doesn't support Bosnian.
\end{itemize}

Table \ref{table:benchmark_results} shows the resulting accuracies.  Although LIDE had lack of  competitive advantage in this unfair test, it surpassed the industry leaders in terms accuracy.

\begin{table}[h]
\centering
\begin{tabular}{@{}lc@{}}
\toprule
Solution & Accuracy \\ \midrule
LIDE & 95\% \\
Google Translate API & 89\% \\
Rosette Language API & 86\% \\
langid.py & 80\% \\
Yandex Translator API & 79\% \\
\bottomrule
\end{tabular}
\vspace{0.2cm}
\caption{Benchmark results in non-increasing order of accuracy}
\label{table:benchmark_results}
\end{table}

\subsubsection*{Comparative test design}
We queried the \textit{test file} of dataset of DSL Shared Task 2015  and compared the resulting predictions with the labels in the \textit{test-gold} file.  We didn't employ any training session with benchmarked solutions, since, these solutions were already trained and claimed to be ready for general purpose use.

None of these solutions were able to distinguish the language varieties.
i.e. 
\begin{itemize}
\item They predicted simply \textit{Spanish} for Ibero-Romance (Spanish) language group.
\item They predicted simply \textit{Portuguese} for Ibero-Romance (Portuguese) language group.
\end{itemize}
Therefore we accepted the prediction of the benchmarked solution if it predicted the main language correctly.  Otherwise we considered it to be a misprediction.

\begin{figure}
\centering
\resizebox{\columnwidth}{!}{
\begin{tabular}{cc}
\includegraphics[width=40mm]{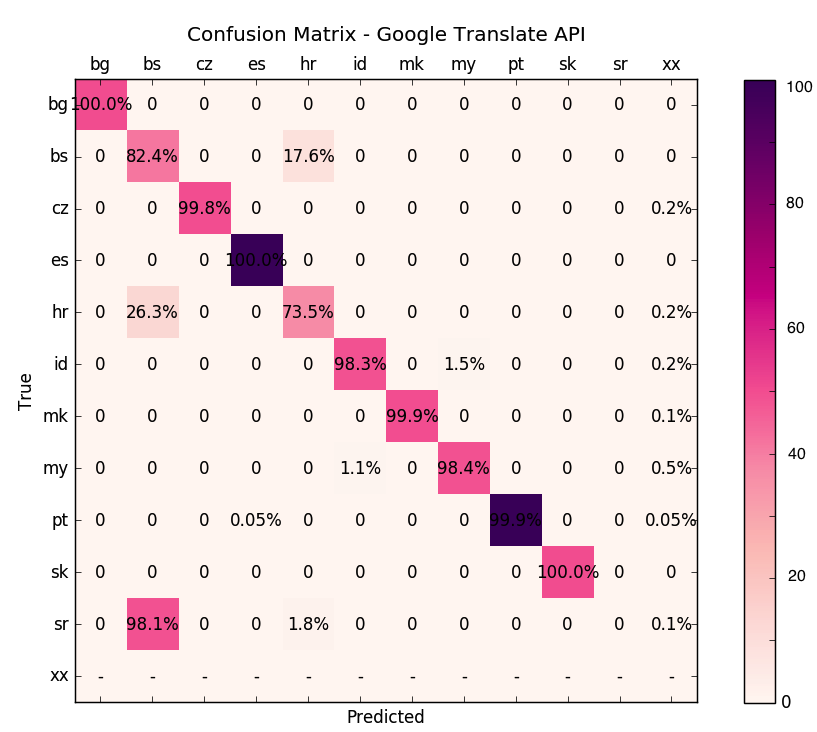}
\label{fig:bench_google}
&
\includegraphics[width=40mm]{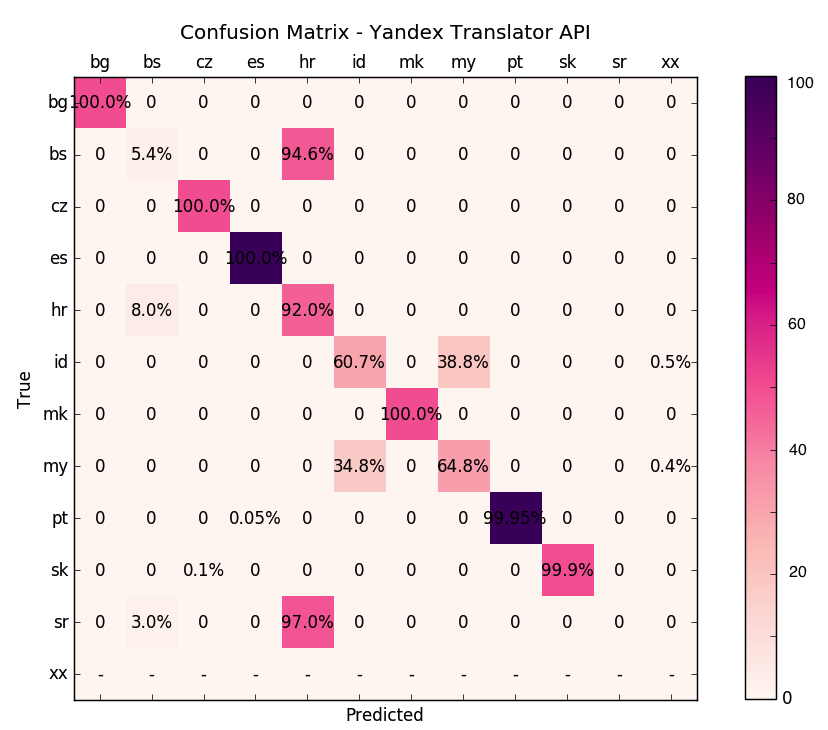}
  \label{fig:bench_yandex}\\
\small (A) Google Translate API &  (B) Yandex Translator API \\
\includegraphics[width=40mm]{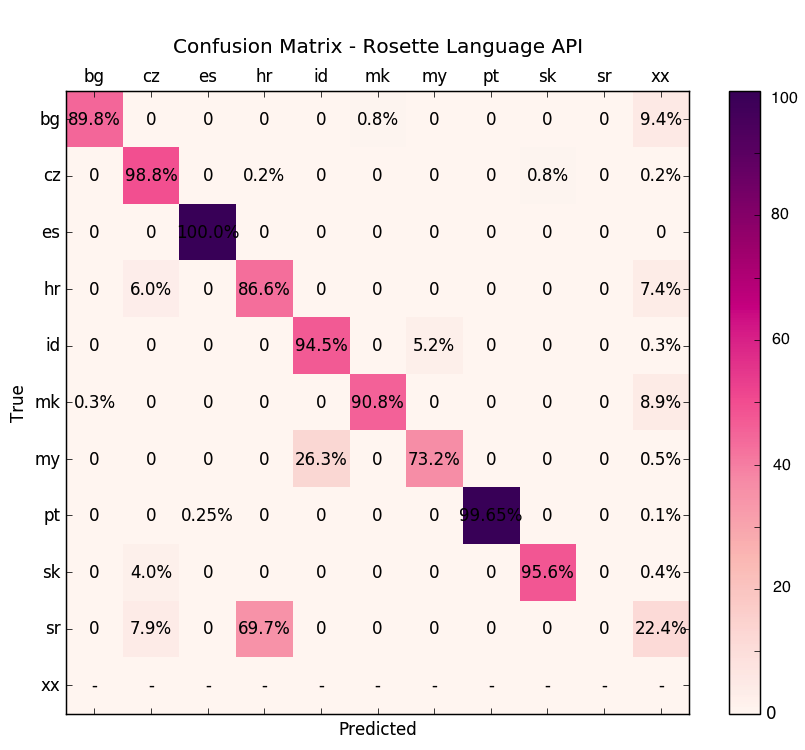}
  \label{fig:bench_rosette}
&
\includegraphics[width=40mm]{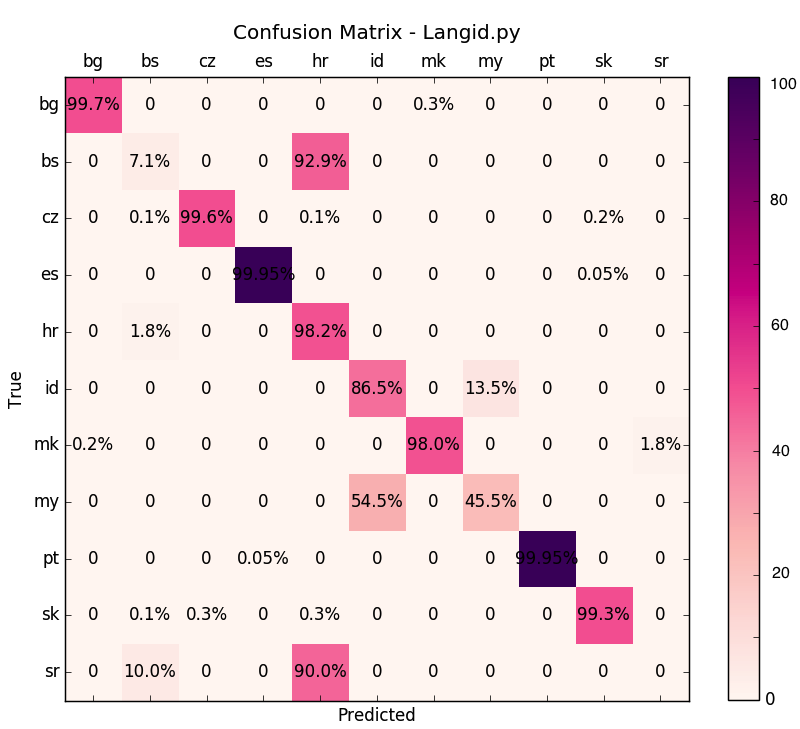}
\label{fig:bench_langid}\\
\small (C) Rosette Language API &  (D) langid.py \\
\end{tabular}
}
\caption{Confusion matrices of benchmarked solutions}
\label{fig:confusion_matrices}
\end{figure}

Table \ref{table:benchmark_results} shows the accuracy and Figure \ref{fig:confusion_matrices}  shows the confusion matrices of each solution.  Below is a comparative analysis  relative to our solution.

\subsection*{Google Translate API \footnote{https://cloud.google.com/translate}}
Google  provides a language detection service for 91 languages. LIDE surpassed in distinguishing South Western Slavic group.  The accuracy of Google Translate API for this particular group was 38\% lower than LIDE which formed the main difference of the overall accuracy between LIDE and Google Translate API.

\subsection*{Yandex Translator API \footnote{https://tech.yandex.com/translate}}
Yandex supports 63 languages.  Similar to Google, Yandex failed to distinguish the languages in South Western Slavic group.  Hence, it moved off one step in the overall accuracy relative to LIDE.  In addition, Yandex showed a very low  accuracy of 62.75\% in Astronesian group which moved off a second step in the overall accuracy relative to LIDE.

\subsection*{Rosette Language API \footnote{https://developer.rosette.com}}
Rosette Language API supports 54 languages excluding Bosnian. Since Bosnian is excluded in the language inventory of Rosette, we discarded Bosnian sentences and queried the remaining languages.  Apart from the Bosnian sentences, Rosette has showed highly similar accuracy  characteristics with Yandex Translator API.

\subsection*{langid.py}
Langid.py is an off-the-shelf language identification tool and it is considered to be a cornerstone in the literature \cite{Lui2012}. langid.py shared very similar accuracy  characteristics with Yandex Translator API, with a subtle difference that langid.py came up with slightly higher accuracy both in Astronesian and South Western Slavic groups. It should be noted that langid.py is the software that is owned and used by one of the competitors of DSL 2014, namely UniMelb-NLP.

\begin{table}[h]
\centering
\begin{tabular}{@{}lc@{}}
\toprule
Solution & Accuracy \\ \midrule
LIDE & 95\% \\
Google Translate API & 89\% \\
Rosette Language API & 86\% \\
langid.py & 80\% \\
Yandex Translator API & 79\% \\
\bottomrule
\end{tabular}
\vspace{0.2cm}
\caption{Benchmark results in non-increasing order of accuracy}
\label{table:benchmark_results}
\end{table}

\section{Conclusion and next steps }

We have presented a deep neural network based language identification scheme that achieves near perfect accuracy in classifying dissimilar languages and about 90\% accuracy on highly similar languages.   
Specifically, the languages in Western Slavic Slavic group posed the highest challenge. And expanding the corpus of these languages using external sources did not help much mainly because no n-grams of words that are unique to certain languages were ingested by the expanded part of the corpus.
We have relied on the ensemble of RNN models to discover the structure unique to a specific language but we could not engineer any additional feature due to lack of knowledge in those specific languages. 
At this point, we think, further improvement can only be achieved by designing rule based features by talking to language experts or native speakers.

\section*{Acknowledgments}
We would like to thank \textit{David Jurgens} from Department of Computer Science, Stanford University for helping with the initial idea, dataset and previous research,  \textit{Junjie Qin} from Department of Computational and Mathematical Engineering, Stanford University for his mentoring and insightful comments that polished the outcome of the study, \textit{AWS Educate Program} for providing EC2 credits and computing resources and 
\textit{Microsoft Azure for Research Program} for providing Azure credits and full featured computing resources, and lastly \textit{Google}, \textit{Yandex} and \textit{Basis Tech} for providing free access to their language detection APIs for our benchmarking analysis.
\vfill

\bibliography{Mendeley,Priyank} 
\bibliographystyle{IEEEtran.bst}
\end{document}